%% file: pess_2026_generalize.tex
\documentclass[conference]{IEEEtran}
\IEEEoverridecommandlockouts
\usepackage{cite}
\usepackage{amsmath,amssymb,amsfonts}
\usepackage{algorithmic}
\usepackage{graphicx}
\usepackage{textcomp}
\usepackage{xcolor}
\usepackage{comment}
\usepackage{stfloats}
\usepackage{tikz}
\usetikzlibrary{
    arrows.meta,
    positioning,
    shapes.geometric,
    patterns
}

\usepackage[nolist]{acronym}
\usepackage{booktabs}

\def\BibTeX{{\rm B\kern-.05em{\sc i\kern-.025em b}\kern-.08em
    T\kern-.1667em\lower.7ex\hbox{E}\kern-.125emX}}

\begin{document}
\input{acronyms}

\title{Characterizing Cross-Relay Generalization Failure in Machine Learning-Based Power System Protection}

\author{
\IEEEauthorblockN{
Nithin Pradeep Nadayil Kizhakkethil\textsuperscript{1,*},
Julian Oelhaf\textsuperscript{1},
Christian Bergler\textsuperscript{2},
Andreas Maier\textsuperscript{1},
Siming Bayer\textsuperscript{1}
}

\IEEEauthorblockA{
\textit{\textsuperscript{1}Pattern Recognition Lab, Friedrich-Alexander-Universit\"at Erlangen-N\"urnberg} \\
\textit{\textsuperscript{2}Department of Electrical Engineering, Media and Computer Science, Ostbayerische Technische Hochschule Amberg-Weiden} \\
{\textsuperscript{*}Corresponding author: nithin.pradeep.nadayil@fau.de}
}
}

\maketitle

\begin{abstract}
Machine learning-based fault detectors are often evaluated at the relay locations used for training, although local voltage and current measurements vary across power networks. This work investigates cross-relay generalization for binary detection of faults anywhere in the network using local measurements, rather than line-selective tripping. Using 50 ms windows from a simulated 110 kV transmission network, we evaluate a multilayer perceptron across all 64 source–target combinations of eight relays and five random seeds. A shared simulation-level split and source-only threshold selection provide a controlled held-out evaluation without target-specific calibration. Despite nearly unchanged mean F1-scores (0.730 within-relay versus 0.734 cross-relay), the mean false-positive rate (FPR) increases from 0.088 to 0.341, while the mean false-negative rate decreases from 0.399 to 0.318. The change is strongly directional: models trained at Bus 2 relays yield a mean FPR of 0.670 at Bus 1 and Bus 3 relays, compared with 0.096 for cross-relay transfers within the same bus group. Linear probing indicates that learned representations retain coarse bus-group information, while exploratory class-conditional latent distribution analysis shows no consistent monotonic association with transfer errors. These findings demonstrate that aggregate metrics can conceal location-structured changes in false-alarm behavior, highlighting the importance of evaluating error types separately when assessing cross-relay generalization.
\end{abstract}


\section{Introduction}
Power system protection is a fundamental component of modern electrical grids, enabling the rapid detection and isolation of faults to maintain system stability, equipment safety, and operational reliability. \Acp{pr} continuously monitor local electrical quantities, such as phase voltages and currents, and initiate protective actions when abnormal operating conditions are detected. With the increasing deployment of digital substations and standardized communication protocols such as IEC 61850, large volumes of high-resolution relay measurements have become available, creating new opportunities for \ac{ml}-based protection systems \cite{aftab_iec_2020,kordowich_simulation_2026}.

\Ac{ml} has emerged as a promising approach for fault detection in power system protection \cite{abdullah_ultrafast_2018,oelhaf_systematic_2025}. Recent studies have investigated \ac{ml}-based methods for fault detection, fault classification, and fault localization using electrical measurements, including \ac{mlp}-based approaches for transmission-line protection \cite{tran_integration_2023,n_transmission_2021}. A broad range of learning architectures has been explored in power system protection, including conventional \ac{ml} methods and deep learning models such as \acp{mlp}, Convolutional Neural Networks (CNNs), Recurrent Neural Networks (RNNs), and topology-aware Graph Neural Networks (GNNs)~\cite{oelhaf_scoping_2025,kordowich_graph_2025}. While these approaches often report strong performance under their respective evaluation settings, their ability to generalize across different measurement locations remains less clearly characterized.

However, such evaluations often implicitly assume comparable measurement distributions across training and deployment locations. In practical power systems, this assumption may not hold. \Acp{pr} are installed at different locations throughout the transmission network, and the electrical signals observed at each relay depend on their position relative to the fault, network topology, and power flow conditions. Consequently, measurements recorded at different relay locations exhibit distinct statistical characteristics, while the asymmetric changes in classification errors under pairwise cross-relay transfer remain insufficiently characterized for relay-local models. In particular, it remains unclear which error type dominates, whether aggregate metrics reveal this behavior, and whether the failure exhibits topology-related structure.

Although generalization under distribution shift has received increasing attention in machine learning~\cite{gulrajani_search_2021,koh_wilds_2021}, relatively little attention has been given to cross-relay generalization in power system protection. Existing work primarily evaluates models using random train-test splits or fixed measurement locations, making it difficult to determine whether the learned representations capture relay-independent fault characteristics or simply memorize relay-specific measurement distributions.

Although topology-aware approaches have demonstrated generalization to unseen grids without retraining \cite{kordowich_graph_2025}, the transfer behavior of models trained using measurements from individual relay locations remains less well characterized. This work therefore investigates pairwise cross-relay transfer using a controlled \ac{mlp} baseline, with particular emphasis on asymmetric changes in false-positive and false-negative rates.

Answering this requires more than overall classification performance: accuracy or F1-score may change little while the types of errors change substantially, and the learned representations may still encode relay-location information. We therefore investigate cross-relay generalization using exhaustive pairwise transfer experiments between all transmission-line \acp{pr}, examining overall performance, false-positive and false-negative behavior, and relay-location structure in the learned representations. The objective is not to assume that transfer should be invariant, but to determine how and to what extent generalization fails, and whether the failures show systematic structure related to relay location and network topology.

\section{System Setup and Data}

\subsection{EMT Simulation Dataset}

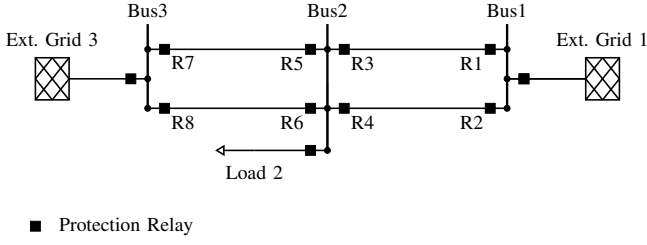
\begin{figure}[t]
    \centering
    \resizebox{\columnwidth}{!}{%
        \input{figures/Fig1}
    }
    \caption{Schematic of the Double Line grid topology. The diagram illustrates the transmission lines, the buses, and \acp{pr}. Measurements are recorded at the \acp{pr} and transmitted to the substations at each bus. }
    \label{fig:grid_topology}
\end{figure}

The experiments in this work are conducted using the AdaptGrid subset of the EvEMTBench electromagnetic transient (EMT) simulation dataset generated in DIgSILENT PowerFactory~\cite{kordowich_simulation_2026}. The dataset is based on a 110~kV double-line transmission network consisting of three buses interconnected by four transmission lines, as illustrated in Fig.~\ref{fig:grid_topology}. The network is instrumented with 11 measurement locations, comprising eight transmission-line \acp{pr}, two external-grid measurement points, and one load measurement point. Each \ac{pr} records synchronized three-phase voltage and current measurements~\cite{kordowich_simulation_2026}.

Each measurement location records three-phase voltages and currents sampled at
9.6~kHz. The six electrical quantities measured at location $m$ are represented as

\begin{equation*}
\mathbf{x}_m(t)=
\begin{bmatrix}
I_A(t) &
I_B(t) &
I_C(t) &
V_A(t) &
V_B(t) &
V_C(t)
\end{bmatrix}^{\top},
\end{equation*}

where $I$ and $V$ denote the three-phase currents and voltages, respectively.
Concatenating the measurements from all eleven locations yields a
66-dimensional feature vector for each simulation time step.

The dataset incorporates significant operating-condition variability through randomization of loading conditions, source characteristics, fault locations, inception angles, and fault impedances. This produces a diverse collection of operating scenarios that enables a comprehensive evaluation of machine learning models under varying network conditions.

\subsection{Fault Scenarios and Class Distribution}

The experiments use fault-event simulations from the AdaptGrid subset of EvEMTBench~\cite{kordowich_simulation_2026}, excluding switching events. The dataset contains 4,667 simulations and 107,341 windows per relay, comprising 73,000 fault (68.01\%) and 34,341 non-fault (31.99\%) windows.

The binary label \texttt{y\_fault\_present} denotes the global network fault state. Empirical verification showed that all labels agree with an event-overlap ratio threshold of $1/240$: windows with \texttt{overlap\_ratio} $\geq 1/240$ are labeled as faults. Positive labels include fault initiation, ongoing faults, and fault clearing, rather than fault onset alone.

The dataset covers seven fault categories, including single-line-to-ground, line-to-line, double-line-to-ground, three-phase, and incipient faults, with varying fault locations and operating conditions. The task therefore concerns global fault-state detection from local relay measurements, not conventional zone-selective tripping.

\subsection{Window Extraction and Preprocessing}

Following the sliding-window preprocessing approach in~\cite{oelhaf_systematic_2025}, each EMT simulation is segmented into fixed-length overlapping windows of synchronized three-phase voltage and current measurements. 

Let

\begin{equation*}
\mathbf{X}=\{\mathbf{x}(t)\}_{t=1}^{T},
\end{equation*}

denote the multivariate signal of a single simulation, where

\begin{equation*}
\mathbf{x}(t)\in\mathbb{R}^{66}.
\end{equation*}

A window beginning at time index $k$ is defined as

\begin{equation*}
\mathbf{W}_k=
\begin{bmatrix}
\mathbf{x}(k)\\
\mathbf{x}(k+1)\\
\vdots\\
\mathbf{x}(k+L-1)
\end{bmatrix}
\in\mathbb{R}^{L\times66},
\end{equation*}

where $L=480$ samples corresponds to a duration of $50\,\mathrm{ms}$ at a sampling frequency of $9.6\,\mathrm{kHz}$. Consecutive windows are extracted every $5\,\mathrm{ms}$, producing overlapping windows that capture the temporal evolution of the fault and provide a new input window every $5\,\mathrm{ms}$.

Each extracted sample is therefore represented as

\begin{equation*}
\mathbf{W}_k\in\mathbb{R}^{480\times66}.
\end{equation*}

Since this work investigates relay generalization, only the eight transmission-line \acp{pr} are considered in the subsequent experiments. For each transmission-line \ac{pr}, indexed by $r\in\{1,\ldots,8\}$, only the six locally measured channels are extracted from the global window $\mathbf{W}_k$, yielding the relay-specific input

\begin{equation}
\mathbf{W}_k^{(r)} \in \mathbb{R}^{480\times6}.
\label{eq:relay_input}
\end{equation}
where $r$ denotes the relay under consideration.

\section{Methods}

This work characterizes cross-relay generalization through three complementary aspects: predictive performance under source-to-target transfer, changes in error behavior (\ac{fpr} and \ac{fnr}), and differences in the learned representations across relay locations, analyzed using linear probes and MMD$^2$. A multilayer perceptron (\ac{mlp}) serves as a controlled baseline to isolate the influence of relay-location-dependent measurement distributions.

\subsection{Multilayer Perceptron}

The \ac{mlp} operates on minimally processed, raw relay-local measurements without input normalization or hand-crafted feature engineering. As defined in (\ref{eq:relay_input}), each input window of raw three-phase voltages and currents is flattened into a feature vector
\begin{equation*}
\mathbf{z}\in\mathbb{R}^{2880}.
\end{equation*}

It is processed by two fully connected hidden layers with 256 and 128 neurons, each followed by a ReLU activation. A final linear output neuron produces a fault logit, converted to a probability by a sigmoid function during inference. The network has approximately 770,561 trainable parameters. As this architecture is a controlled baseline rather than a representative protection model, the magnitude of the observed degradation should not be generalized to other architectures.

\subsection{Training Settings}

The \ac{mlp} is trained with binary cross-entropy with logits and the Adam optimizer (learning rate $1.38\times10^{-4}$, no weight decay), using a batch size of 1024 for at most 200 epochs. Early stopping with a patience of 20 epochs is applied, and the parameters from the best validation-loss epoch are retained.

After training, the operating threshold is selected on the source relay's validation set by maximizing Youden's $J$ statistic,

\begin{equation*}
J(t)=\mathrm{TPR}(t)-\mathrm{FPR}(t).
\end{equation*}

The threshold is then frozen and used for the source relay and all target relays, so no target-relay information influences the operating point. ROC AUC is computed from continuous raw model logits and is independent of this threshold.

Performance is reported as accuracy, F1-score, ROC AUC, \ac{fpr}, and \ac{fnr}, with the fault class as positive. \ac{fpr} is the fraction of non-fault windows classified as faults, and \ac{fnr} is the fraction of fault windows classified as non-faults.

\subsection{Pairwise Cross-Relay Evaluation}

Each of the eight transmission-line protection relays serves as a source for an independent model, which is evaluated at the source relay and at each of the seven target relays. This yields an $8\times8$ transfer matrix of 64 pairs: eight self evaluations on the diagonal and 56 ordered cross-relay evaluations off the diagonal.

A shared simulation-level split, generated with seed 42,
is reused across all eight relays. It comprises 3,266
training, 700 validation, and 701 test simulations,
corresponding to 75,118, 16,100, and 16,123 windows
per relay, respectively. All windows originating from
the same simulation remain in the same partition.
Self- and cross-relay evaluations use the same
held-out test simulations.

The complete experiment is repeated for five random seeds (42, 5, 10, 25, and 100), which affect model initialization and training but not the fixed simulation-level partition. Results are reported as mean $\pm$ standard deviation across seeds.

To examine location structure, the 56 ordered cross-relay pairs are divided by bus group (Bus~1: R1, R2; Bus~2: R3--R6; Bus~3: R7, R8) into within-group pairs (16), Bus~1--Bus~3 pairs (8), pairs from Bus~2 to Bus~1 or Bus~3 (16), and pairs from Bus~1 or Bus~3 to Bus~2 (16). Mean \ac{fpr} and \ac{fnr} are computed per category from the five-seed mean transfer matrices.

\subsection{Representation Analysis}

Representations are taken from the 128-dimensional penultimate layer of each source-trained \ac{mlp} of the seed-42 run. All representation analyses use one fixed study subset, drawn once and embedded by every source-trained model, so that differences between models are not caused by different input samples. For each of the eight relays, 150 fault and 150 non-fault windows are sampled uniformly at random without replacement with seed 42 from all non-switching windows of that relay, giving 300 windows per relay and 2,400 in total. The subset is not restricted to a data partition and therefore contains windows from training, validation, and test simulations; it comprises 1,878 distinct simulations. Sampling is class-balanced and performed independently for each relay, so relays are not matched on simulation or window position.

Relay identity is predicted from the representations with multinomial logistic regression ($\ell_2$ penalty, $C=1$) on standardized features, and bus-group identity (Bus~1, Bus~2, or Bus~3) with a separate probe of the same type. Because overlapping windows from one simulation are strongly dependent, both probes are evaluated with five-fold stratified cross-validation grouped by simulation, so that all windows from the same EMT simulation fall in the same fold. The chance level is $1/8$ for relay identification. Because Bus~2 contains four relays and Bus~1 and Bus~3 two each, the majority-class baseline for bus-group identification is 0.50. To quantify the relay-identification result, the domain confusion score is computed from the mean out-of-fold accuracy $A_{\text{domain}}$ and $A_{\text{chance}}=1/8$:

\begin{equation}
C = 1 - \frac{A_{\text{domain}} - A_{\text{chance}}}{1 - A_{\text{chance}}},
\label{eq:confusion}
\end{equation}

where $C=1$ corresponds to chance-level relay identification and $C=0$ to perfect identification.

Differences between source and target representations are quantified with the squared maximum mean discrepancy (MMD$^2$)~\cite{gretton_kernel_2012}. Let $\{\mathbf{z}_i^{s}\}_{i=1}^{n}$ and $\{\mathbf{z}_j^{t}\}_{j=1}^{m}$ denote the source- and target-relay representations. The empirical estimator is given in~\eqref{eq:mmd}.

\begin{equation}
\begin{aligned}
\widehat{\mathrm{MMD}}^2
={}&
\frac{1}{n^2}\sum_{i=1}^{n}\sum_{i'=1}^{n}
k(\mathbf{z}_i^{s},\mathbf{z}_{i'}^{s}) \\
&+
\frac{1}{m^2}\sum_{j=1}^{m}\sum_{j'=1}^{m}
k(\mathbf{z}_j^{t},\mathbf{z}_{j'}^{t}) \\
&-
\frac{2}{nm}\sum_{i=1}^{n}\sum_{j=1}^{m}
k(\mathbf{z}_i^{s},\mathbf{z}_j^{t}),
\end{aligned}
\label{eq:mmd}
\end{equation}

with the RBF kernel $k(\mathbf{z},\mathbf{z}')=\exp(-\|\mathbf{z}-\mathbf{z}'\|_2^2/(2\sigma^2))$. The bandwidth is set per source model by the median heuristic on the 300 source-relay embeddings and kept fixed for all target relays of that source. MMD$^2$ is computed separately per class with $n=m=150$: between non-fault windows, which is related to the cross-relay \ac{fpr}, and between fault windows, which is related to the cross-relay \ac{fnr}. This gives 56 ordered source-target pairs per class. The associations with the seed-42 error rates are quantified with Pearson (linear) and Spearman (monotonic) correlation and are interpreted descriptively.


\section{Results}

\subsection{Overall Cross-Relay Performance}

Table~\ref{tab:aggregate_results} summarizes self- and cross-relay performance across five seeds. Cross-relay transfer reduces accuracy (0.7010 to 0.6747) and ROC AUC (0.7929 to 0.7493). The change in error profile is larger: mean \ac{fpr} rises from 0.0879 to 0.3406, while mean \ac{fnr} falls from 0.3986 to 0.3181. Cross-relay degradation is therefore not a uniform increase in errors but a shift towards false positives. Because the threshold is fixed on the source validation set, this shift reflects the model's behavior under the target measurement distribution, not target-specific re-tuning.

The F1-score is nearly unchanged (0.7302 to 0.7336) because the rise in false positives is offset by the fall in false negatives. Aggregate metrics such as F1 can therefore hide substantial changes in error behavior.

Averages also conceal severe individual failures. The worst source-target pair in each seed averages an \ac{fpr} of 0.9081 across seeds. The worst-pair \ac{fnr} (0.5727) exceeds the mean self \ac{fnr}, so the decrease in \ac{fnr} holds on average but not for every pair.

\begin{table}[t]
\centering
\caption{Self-relay, cross-relay, and worst-pair performance across five seeds (mean $\pm$ standard deviation). Worst Cross denotes the worst individual source-target pair for each metric in each seed.}
\label{tab:aggregate_results}
\renewcommand{\arraystretch}{1.2}
\begin{tabular}{lccc}
\toprule
\textbf{Metric} & \textbf{Self} & \textbf{Cross} & \textbf{Worst Cross} \\
\midrule
Accuracy & $0.7010\pm0.0069$ & $0.6747\pm0.0059$ & $0.5734\pm0.0232$ \\
F1       & $0.7302\pm0.0095$ & $0.7336\pm0.0098$ & $0.5747\pm0.0441$ \\
ROC AUC  & $0.7929\pm0.0058$ & $0.7493\pm0.0070$ & $0.6809\pm0.0052$ \\
FPR      & $0.0879\pm0.0134$ & $0.3406\pm0.0379$ & $0.9081\pm0.0908$ \\
FNR      & $0.3986\pm0.0159$ & $0.3181\pm0.0256$ & $0.5727\pm0.0568$ \\
\bottomrule
\end{tabular}
\end{table}

\subsection{Error Behavior Across Relay Pairs}

\begin{figure*}[t]
    \centering
    \includegraphics[width=0.49\textwidth]{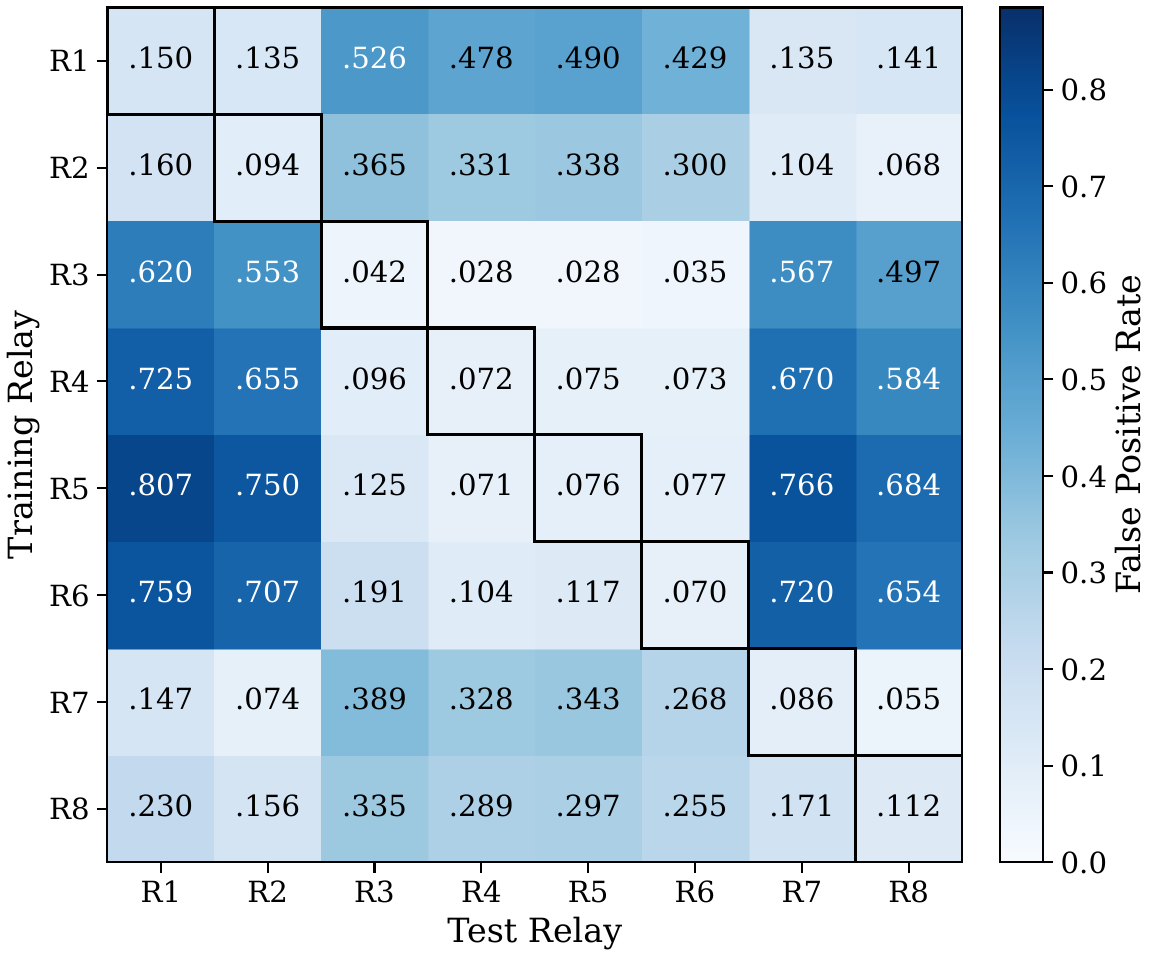}
    \hfill
    \includegraphics[width=0.49\textwidth]{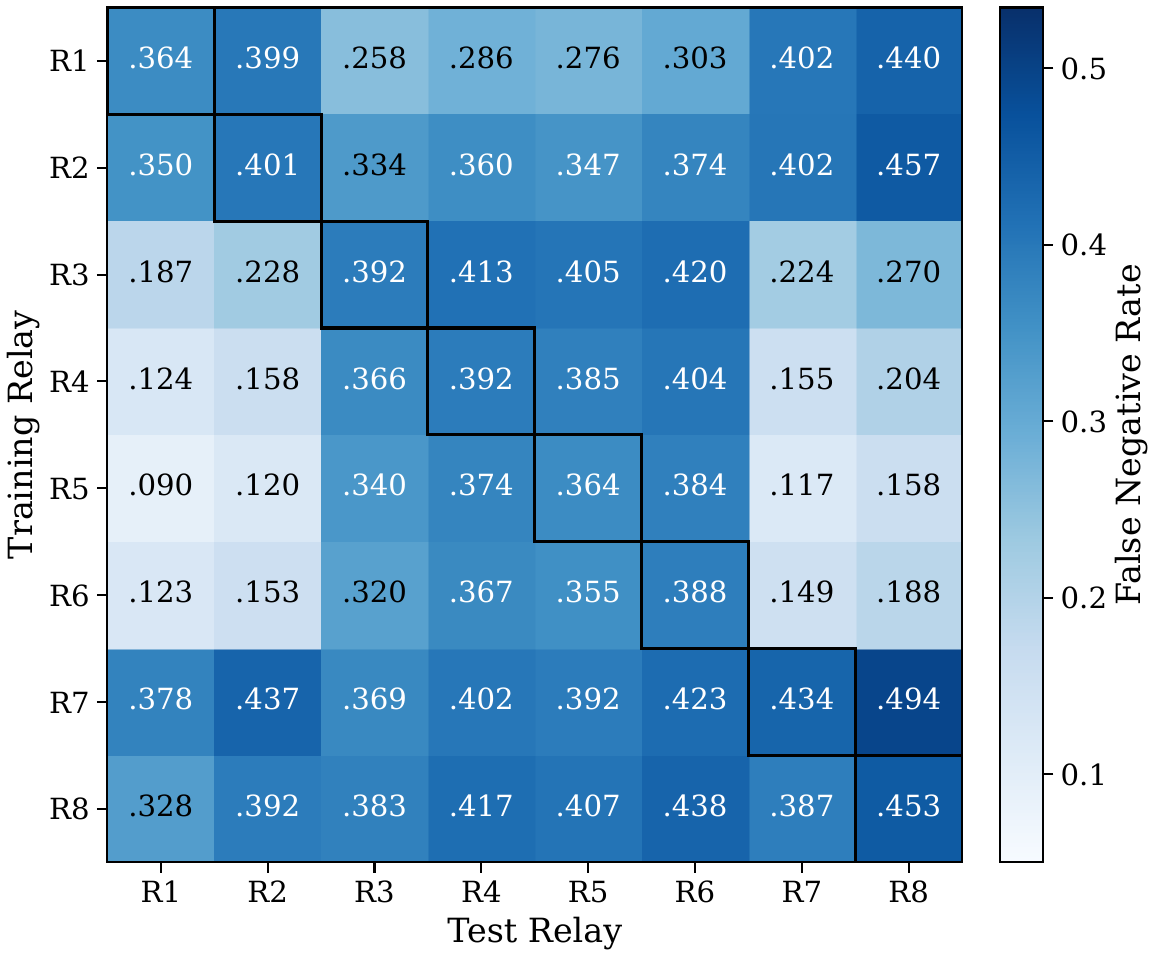}
    \caption{Pairwise cross-relay transfer matrices for the MLP baseline (mean over five seeds). Rows denote the training relay and columns the test relay; diagonal entries are within-relay evaluation. \textit{(a)}~\ac{fpr}. \textit{(b)}~\ac{fnr}.}
    \label{fig:heatmaps}
\end{figure*}

Fig.~\ref{fig:heatmaps} shows the pairwise \ac{fpr} and \ac{fnr} matrices, averaged over five seeds. False-positive behavior depends strongly on the source-target combination. For models trained at Bus~2 relays, the mean \ac{fpr} at Bus~1 and Bus~3 relays ranges from 0.497 to 0.807, whereas transfer within a bus group and between Bus~1 and Bus~3 remains close to the self-relay level. The \ac{fnr} matrix shows a different pattern, and pairs with high \ac{fpr} do not have high \ac{fnr}. Together with the aggregate results, this indicates that the main deterioration under cross-relay transfer is the increase in false-positive rate.

The degradation is also structured by location. Table~\ref{tab:bus_groups} quantifies this structure. Models trained at Bus~2 relays show a mean \ac{fpr} of 0.670 at Bus~1 and Bus~3 relays. Transfer within a bus group (0.096) and between Bus~1 and Bus~3 (0.132) stays close to the mean self-relay level of 0.088. The reverse direction, from Bus~1 or Bus~3 to Bus~2, is elevated but weaker (0.360). The \ac{fnr} shows the complementary behavior: Bus~2-trained models rarely miss faults at Bus~1 and Bus~3 (0.166), consistent with their tendency to flag windows there as faulty. Source relays with the lowest mean cross-relay \ac{fpr} (R2, R7, R8) also show the highest mean cross-relay \ac{fnr}, which indicates a more conservative operating point and not better transfer. Cross-relay failure is therefore not a random loss of performance between arbitrary relays. The task uses the global fault-state label from local measurements, so these results characterize cross-location generalization and not conventional zone-selective relay operation.

\begin{table}[t]
\centering
\caption{Mean cross-relay \ac{fpr} and \ac{fnr} by bus-group relationship of source and target relay. Values are averages over the relay pairs in each category, computed from the five-seed mean transfer matrices (Fig.~\ref{fig:heatmaps}).}
\label{tab:bus_groups}
\renewcommand{\arraystretch}{1.2}
\begin{tabular}{lccc}
\toprule
\textbf{Source $\rightarrow$ target} & \textbf{Pairs} & \textbf{FPR} & \textbf{FNR} \\
\midrule
Within bus group              & 16 & 0.096 & 0.385 \\
Bus 1 $\leftrightarrow$ Bus 3 & 8  & 0.132 & 0.405 \\
Bus 2 $\rightarrow$ Bus 1/3   & 16 & 0.670 & 0.166 \\
Bus 1/3 $\rightarrow$ Bus 2   & 16 & 0.360 & 0.361 \\
\bottomrule
\end{tabular}
\end{table}

\subsection{Representation Analysis}

On the class-balanced study subset, individual relay identity is only weakly recoverable from the penultimate-layer representation under simulation-grouped cross-validation. The mean out-of-fold accuracy across the eight source-trained models is 0.266, compared with a chance level of 0.125. Applying (\ref{eq:confusion}) gives

\begin{equation*}
C = 1 - \frac{0.266 - 0.125}{1 - 0.125} = 0.839.
\end{equation*}

The representation is therefore not strongly discriminative at the individual-relay level, but it is also not fully relay-invariant.

\begin{figure}[t]
\centering
\includegraphics[width=\columnwidth]{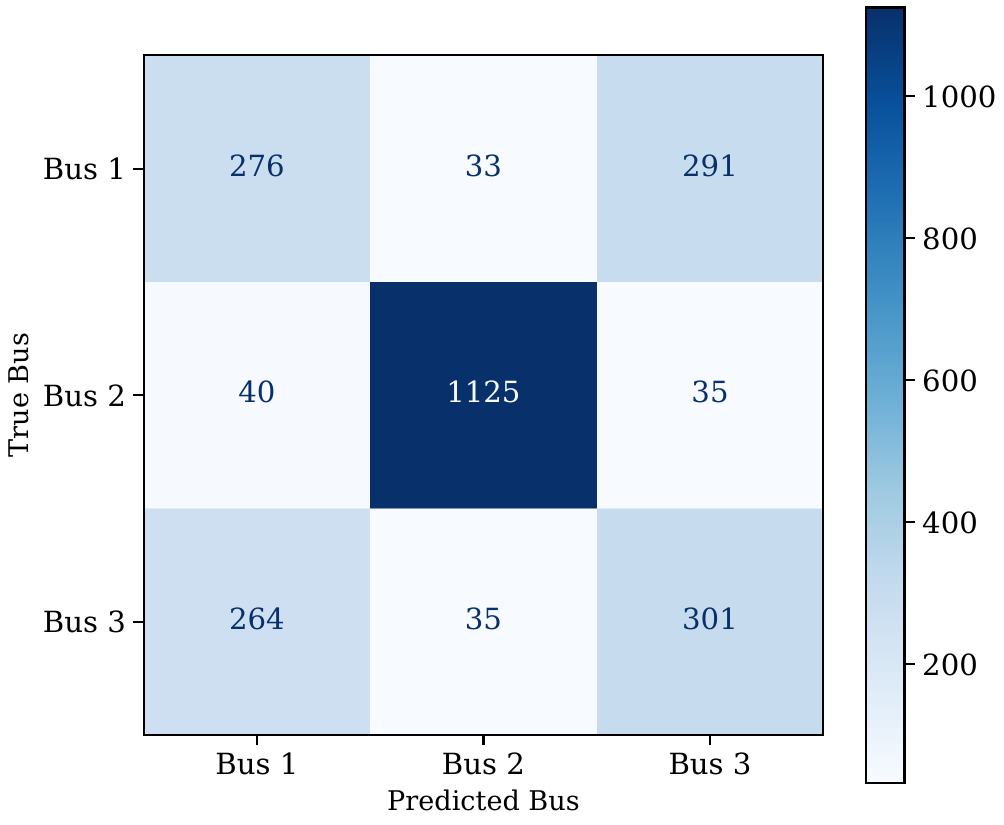}
\caption{Linear probe confusion matrix for bus-group identification (Bus~1, Bus~2, Bus~3) from the penultimate-layer activations of the MLP trained on relay~R1, shown as an illustrative example (grouped five-fold cross-validation, out-of-fold predictions). The confusion pattern shows that coarse bus-group information remains in the representation.}
\label{fig:linear_probe}
\end{figure}

Coarser location information is more recoverable. The bus-group probe reaches a mean out-of-fold accuracy of 0.709 across the eight models, compared with 0.50 for always predicting Bus~2. The confusion matrix in Fig.~\ref{fig:linear_probe} shows that Bus~1 measurements are mostly predicted as Bus~1 or Bus~3, and Bus~2 measurements are mostly predicted as Bus~2. This coarse structure is consistent with the transfer pattern: Bus~1 and Bus~3 measurements are confused with each other, and transfer between Bus~1 and Bus~3 relays stays close to the self-relay \ac{fpr} (0.132), whereas Bus~2 measurements are separated from both and Bus~2-trained models fail at Bus~1 and Bus~3 relays (\ac{fpr} 0.670). The probe is symmetric and cannot explain why the reverse direction is weaker. The analysis is diagnostic, not causal: retained location information does not by itself show that it causes the degradation.

\subsection{Representation Distribution Differences and Error}

MMD$^2$ is computed per class and related to the matching error rate over the 56 ordered pairs of the seed-42 run. Between non-fault windows, MMD$^2$ has a Pearson correlation of $r=-0.384$ and a Spearman correlation of $\rho=-0.190$ with cross-relay \ac{fpr}. Between fault windows, the corresponding values with \ac{fnr} are $r=0.349$ and $\rho=0.045$. The \ac{fpr} association is negative, so larger latent distances are associated with lower false-positive rates, opposite to the naive expectation that greater shift means worse transfer. The rank correlations are weak, so the linear associations do not indicate a robust monotonic relationship.

The observed MMD$^2$ values indicate differences between relay-specific latent distributions, but show no consistent monotonic association with cross-relay transfer errors. Source and target samples with similar aggregate distributions can still lie differently relative to the decision boundary, so MMD$^2$ is best treated as a diagnostic of representation shift and not as an explanation of generalization failure. A possible explanation for the negative \ac{fpr} association is that a source model that flags most non-fault target windows as faults maps them into a compact region of its latent space, which can reduce the measured MMD$^2$ even though the transfer error is large. This hypothesis is not tested here. The correlations come from a single seed, the 56 ordered pairs share source models and target relays and are therefore not independent, and the bandwidth is source-specific, so MMD$^2$ values are not strictly comparable across sources. The correlations are therefore reported descriptively.

\section{Conclusion}

This work characterized how an \ac{mlp} fault detector trained at one relay location behaves at other locations, using exhaustive pairwise transfer across eight transmission-line \acp{pr} in a simulated 110~kV network, with a shared simulation-level split and a source-validation threshold. Overall degradation is moderate: mean accuracy falls from 0.7010 to 0.6747 and ROC AUC from 0.7929 to 0.7493. The error profile changes far more. Mean \ac{fpr} rises from 0.0879 to 0.3406, with the per-seed worst relay pair averaging 0.908, while mean \ac{fnr} falls from 0.3986 to 0.3181, so the F1-score is nearly unchanged (0.7302 to 0.7336). Cross-relay transfer thus shifts the balance towards false positives, and aggregate metrics can hide this. The \ac{fnr} decrease holds only on average, since the worst-pair \ac{fnr} (0.573) exceeds the mean self value.

The degradation is structured by location: it depends on the source-target pair, and high \ac{fpr} occurs mainly when models trained at Bus~2 relays are applied at Bus~1 or Bus~3 relays. It is weaker in the reverse direction and stays close to the self-relay level between Bus~1 and Bus~3. Linear probing shows that individual relay identity is only weakly recoverable, but coarse bus-group information remains in the representation, consistent with this pattern. In a single-seed, descriptive analysis, class-conditional MMD$^2$ correlates negatively with \ac{fpr} ($r=-0.38$, $\rho=-0.19$) and positively with \ac{fnr} ($r=0.35$, $\rho=0.04$). The \ac{fpr} sign is opposite to the naive expectation that greater shift means worse transfer, and the rank correlations are weak. MMD$^2$ is therefore a diagnostic of representation shift, not an explanation of transfer error.

These results motivate evaluating cross-relay generalization with \ac{fpr} and \ac{fnr} reported separately. The study is limited to one simulated network, one baseline architecture, and a global fault-state label, so the magnitudes should not be generalized. Future work should test whether domain adaptation and invariant representation learning~\cite{ganin_domain-adversarial_2016} reduce the location dependence, and validate on additional network configurations and measured field data.

\section*{Acknowledgment}
This project was funded by the Deutsche Forschungsgemeinschaft (DFG, German Research Foundation) - 535389056.

\vspace{0.4cm}
\noindent The authors gratefully acknowledge the scientific support and high-performance computing resources provided by the Erlangen National High Performance Computing Center (NHR@FAU) at Friedrich-Alexander-Universit\"at Erlangen-N\"urnberg (FAU) under NHR project b304dc. NHR funding is provided by the German federal and Bavarian state authorities.

\vfill
\pagebreak

\bibliographystyle{IEEEtran}
\bibliography{references}

\end{document}

%% file: acronyms.tex
\newacro{pr}[PR]{protection relay}
\newacro{ml}[ML]{machine learning}
\newacro{mlp}[MLP]{multilayer perceptron}
\newacro{emt}[EMT]{electromagnetic transient}
\newacro{fpr}[FPR]{False Positive Rate}
\newacro{fnr}[FNR]{False Negative Rate}
\newacro{auc}[AUC]{Area Under Curve}
\newacro{mmd}[MMD]{Maximum Mean Discrepancy}

%% file: figures/Fig1.tex
\begin{tikzpicture}[
    x=1cm,
    y=0.75cm,
    relay/.style={draw, fill=black, minimum size=5pt, inner sep=0pt},
    bus/.style={draw, very thick, minimum width=8pt, minimum height=30pt, inner sep=0pt},
    >=Stealth
]

\draw[very thick] (1.8, 1.2) -- (1.8, 3.2);
\node[above] at (1.8, 3.2) {\normalsize Bus3};

\draw[very thick] (5.0, 0.15) -- (5.0,3.2);
\node[above] at (5.0, 3.2) {\normalsize Bus2};

\node[above] at (8.2, 3.2) {\normalsize Bus1};
\draw[very thick] (8.2, 1.2) -- (8.2,3.2);

\draw[thick] (5.0, 2.6) -- (8.2, 2.6);

\draw[thick] (5.0, 1.2) -- (8.2, 1.2);

\draw[thick] (1.8, 2.6) -- (5.0, 2.6);

\draw[thick] (1.8, 1.2) -- (5.0, 1.2);

\node[relay] (R3) at (5.3, 2.6) {};  \node[below right]  at (R3) {\normalsize R3};
\node[relay] (R1) at (7.9, 2.6) {};  \node[below left] at (R1) {\normalsize R1};

\node[relay] (R4) at (5.3, 1.2) {};  \node[below right]  at (R4) {\normalsize R4};
\node[relay] (R2) at (7.9, 1.2) {};  \node[below left] at (R2) {\normalsize R2};

\node[relay] (R7) at (2.1, 2.6) {};  \node[below right]  at (R7) {\normalsize R7};
\node[relay] (R5) at (4.7, 2.6) {};  \node[below left] at (R5) {\normalsize R5};

\node[relay] (R8) at (2.1, 1.2) {};  \node[below right]  at (R8) {\normalsize R8};
\node[relay] (R6) at (4.7, 1.2) {};  \node[below left] at (R6) {\normalsize R6};

\filldraw[fill=black] (1.8, 2.6) circle (1.5pt);
\filldraw[fill=black] (1.8, 1.2) circle (1.5pt);
\filldraw[fill=black] (5.0, 2.6) circle (1.5pt);
\filldraw[fill=black] (5.0, 1.2) circle (1.5pt);
\filldraw[fill=black] (8.2, 2.6) circle (1.5pt);
\filldraw[fill=black] (8.2, 1.2) circle (1.5pt);

\draw[thick] (9.6, 1.4) rectangle (10.2, 2.4);
\draw[thick] (9.6, 1.4) -- (10.2, 2.4);
\draw[thick] (9.6, 2.4) -- (10.2, 1.4);
\draw[thick] (9.6, 1.9) -- (9.9, 2.4);
\draw[thick] (9.6, 1.9) -- (9.9, 1.4);
\draw[thick] (9.9, 2.4) -- (10.2, 1.9);
\draw[thick] (10.2, 1.9) -- (9.9, 1.4);
\node[above] at (9.9, 2.5) {\normalsize Ext. Grid 1};
\filldraw[fill=black] (8.2, 1.9) circle (1.5pt);

\draw[thick] (8.2, 1.9) -- (9.6, 1.9);
\node[relay] at (8.5, 1.9) {};

\draw[thick] (-0.2, 1.4) rectangle (0.4, 2.4);
\draw[thick] (-0.2, 1.4) -- (0.4, 2.4);
\draw[thick] (-0.2, 2.4) -- (0.4, 1.4);
\draw[thick] (-0.2, 1.9) -- (0.1, 2.4);
\draw[thick] (-0.2, 1.9) -- (0.1, 1.4);
\draw[thick] (0.1, 2.4) -- (0.4, 1.9);
\draw[thick] (0.4, 1.9) -- (0.1, 1.4);
\node[above] at (0.1, 2.5) {\normalsize Ext. Grid 3};

\draw[thick] (8.2, 1.9) -- (9.6, 1.9);
\node[relay] at (8.5, 1.9) {};

\node[relay] at (1.5, 1.9) {};
\draw[thick] (1.8, 1.9) -- (0.4, 1.9);
\filldraw[fill=black] (1.8, 1.9) circle (1.5pt);

\node[relay] at (4.7, 0.2) {};
\draw[thick] (5.0, 1.8) -- (5.0, 0.2) -- (3.7, 0.2);
\filldraw[fill=black] (5.0, 0.2) circle (1.5pt);
\draw[thick, -{Triangle[open]}] (3.7, 0.2) -- (3.0, 0.2);
\node[below] at (3.7, 0.0) {\normalsize Load 2};

\node[relay] at (-0.2, -1.6) {};
\node[right] at (0.1, -1.6) {\normalsize Protection Relay};

\end{tikzpicture}